\documentclass[letterpaper, 10 pt, conference]{ieeeconf}  

\IEEEoverridecommandlockouts  
\usepackage[pdftex]{graphicx}
\DeclareGraphicsExtensions{.pdf,.jpeg,.png}
\usepackage{amsmath} 
\usepackage{amssymb}  
\DeclareMathOperator{\E}{\mathbb{E}}
\DeclareMathOperator{\D}{\mathbb{D}}

\usepackage{tabularx,booktabs,multirow}
\usepackage{array}
\usepackage{float}
\usepackage{makecell}
\usepackage{tablefootnote}
\usepackage{multicol}
\usepackage{booktabs}
\usepackage{threeparttable}
\usepackage{url}
\usepackage[font=small,skip=10pt]{caption}
\usepackage[dvipsnames]{xcolor}
\usepackage{amsmath}
\usepackage{xcolor}
\usepackage{algorithmic}
\usepackage{dirtytalk}
\usepackage[ruled,vlined]{algorithm2e}

\newcolumntype{P}[1]{>{\centering\arraybackslash}p{#1}}
\SetKwInOut{Input}{input}
\SetKwInOut{Output}{output}
\makeatletter
\let\NAT@parse\undefined
\makeatother
\usepackage{hyperref}
\hypersetup{colorlinks,allcolors=red}

\title{\LARGE \bf
Vision Based Adaptation to Kernelized Synergies for Human Inspired Robotic Manipulation 
}

\author{Sunny Katyara$^{1,2}$, Fanny Ficuciello$^{2}$,~\IEEEmembership{Senior Member,~IEEE}, Fei Chen$^{1}$,~\IEEEmembership{Member,~IEEE}, \\
Bruno Siciliano$^{2}$,~\IEEEmembership{Fellow,~IEEE},  Darwin G. Caldwell$^{1}$,~\IEEEmembership{Senior Member,~IEEE}, 
\thanks{This research is supported by the project ``Improving Reproducibility in Learning Physical Manipulation Skills with Simulators Using Realistic Variations'' funded by EU H2020 ERA-Net Chist-Era program. \textit{(Corresponding author: Fei Chen)} }
\thanks{$^{1}$ Sunny Katyara, Darwin Caldwell, Fei Chen are with APRIL Lab, Department of Advanced Robotics, Istituto Italiano di Tecnologia, Via Morego 30, 16163, Genova, Italy (e-mail: {\tt\small name.surname@iit.it}).}
\thanks{$^{2}$ Fanny Ficuciello, Bruno Siciliano are with Department of Information Technology and Electrical Engineering and the Interdepartmental Center for Advances in Robotic Surgery, University of Naples Federico II, Naples 80125, Italy ({e-mail: \tt\small name.surname@unina.it}).}
}

\begin{document}

\maketitle
\thispagestyle{empty}
\pagestyle{empty}

\begin{abstract}

Humans in contrast to robots are excellent in performing fine manipulation tasks owing to their remarkable dexterity and sensorimotor organization. Enabling robots to acquire such capabilities, necessitates a framework that not only replicates the human behaviour but also integrates the multi-sensory information for autonomous object interaction. To address such limitations, this research proposes to augment the previously developed kernelized synergies framework with visual perception to automatically adapt to the unknown objects. The kernelized synergies, inspired from humans, retain the same reduced subspace for object grasping and manipulation. To detect object in the scene, a simplified perception pipeline is used that leverages the RANSAC algorithm with Euclidean clustering and SVM for object segmentation and recognition respectively. Further, the comparative analysis of kernelized synergies with other state of art approaches is made to confirm their flexibility and effectiveness on the robotic manipulation tasks. The experiments conducted on the robot hand confirm the robustness of modified kernelized synergies framework against the uncertainties related to the perception of environment.   

\end{abstract}

\section{INTRODUCTION}
\textbf{Robotic manipulation} refers to the ways a robot perceives and interacts with the objects in the environment. The knowledge of the environment i.e, the type and location of objects in the world is a deterministic aspect to improve the performance of robots to manipulate, \textbf{similar to humans}. Building robot hands that emulate the functionalities of humans is challenging owing to the limitations of existing software and hardware \cite{c1}. However, few attempts have been made to mimic the design and control of human hand for flexible grasping and manipulation \cite{c2}\cite{c3}\cite{c4} but are still lagging behind the human capabilities. The control architecture of humans reveals that they exploit reduced subspace, called postural synergies \cite{c5} \cite{c6} to plan and regulate their whole body movements. Inspired from postural synergies, a framework called \textbf{kernelized synergies} has been developed that reuses the same subspace for robust grasping and dexterous manipulation \cite{c7}.

   \begin{figure}[t]
      \centering
      \includegraphics[width=8.5cm]{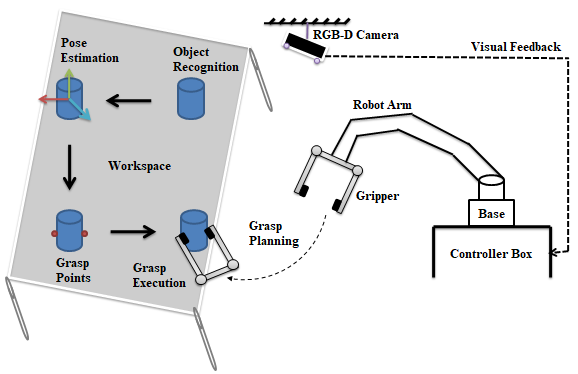}
      \caption{Sketch of vision based adaptation to robotic manipulation. Camera detects the object in the scene and its pose is estimated which is given to controller box for grasp planning and execution.}
      \label{representation}
      \vspace{-15pt}
   \end{figure}

The kernelized synergies framework assumes the pose of objects in the scene to be known for grasp execution. But the human inspired robotic manipulation requires to recognize and estimate the object pose in run time using \textbf{visual perception} as illustrated in Fig. \ref{representation}. The four key steps to the \textit{vision based robotic manipulation} are (1) bringing the gripper within the desirable work-space (2) recognizing and estimating the object pose (3) planning a suitable candidate grasp (4) executing the grasping and manipulation actions. The kernelized synergies framework however considers all the major steps except the object recognition and pose estimation.

The object recognition and pose estimation techniques are still evolving and the algorithms ranging from object semantic segmentation to the end-end learning have been exercised for various robotic tasks \cite{c8}\cite{c9}\cite{c10}. A model based object recognition framework is proposed in \cite{c11} for robotic manipulation. The framework is robust against scalability, complexity and latency of the scene but requires multi-view instances for object recognition. A modified object detection approach based on single image segmentation using discriminative classifier is suggested in \cite{c12} for robotic grasping. Such technique requires the large data base for object reconstruction and does not work well in the cluttered environments. To alleviate the need of building data base and performing well in the complex scenarios, an object detection framework is discussed in \cite{c13} for robotic grasping. The framework however uses model approximations on the filtered point cloud but does not achieve desired accuracy since only two models i.e, spherical and cylindrical are considered for approximation. In our research, we update this pipeline by mitigating the need of using two specific RANSAC models for approximations and proposes to exploit additionally a SVM classifier for object recognition.  

With reference to \cite{c7}, the key contributions of this work are \textbf{(1)} to upgrade the kernelized synergies framework with visual perception for autonomous grasping and manipulation \textbf{(2)} to compare the performance of kernelized synergies framework with other state of art methods using two metrics i.e, normalized square error and primitive accuracy \textbf{(3)} to apply modified framework to perform three different daily life tasks i.e, mounting bulb on the socket, squeezing lemon into the water, and spraying cleanser on the board.

\section{RESEARCH METHODOLOGY}

   \begin{figure}[t]
      \centering
      \includegraphics[width=8.5cm]{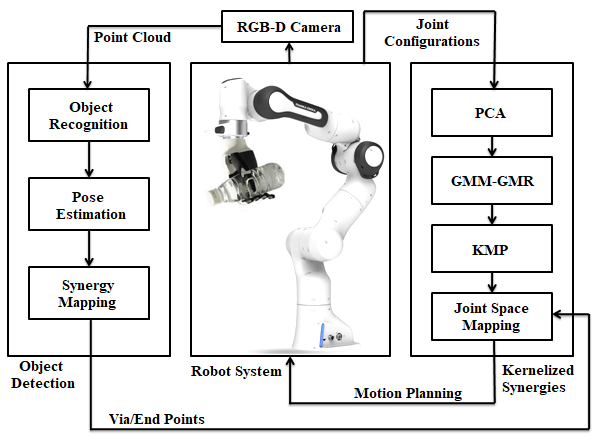}
      \caption{Block diagram of updated kernelized synergies framework.}
      \label{methodology}
      \vspace{-15pt}
   \end{figure}

\begin{figure}[t]
      \centering
      \includegraphics[width=8.5cm]{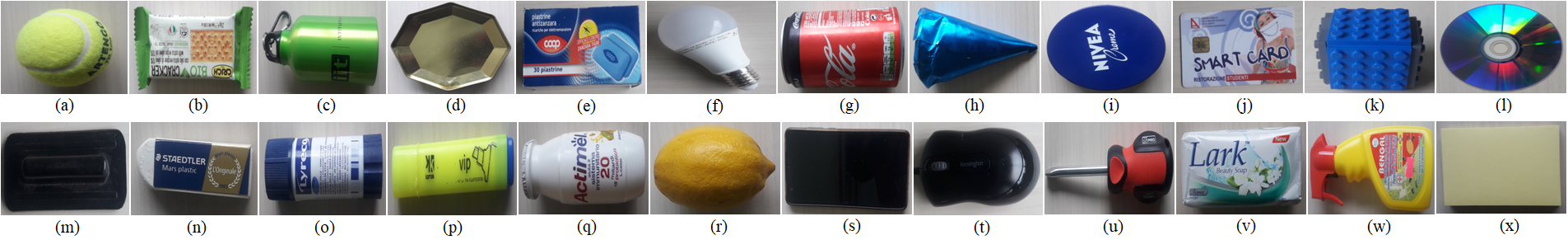}
      \caption{Training data set, (a) ball, (b) biscuit (c) bottle (d) bowl, (e) box, (f) bulb, (g) coke can, (h) cone, (i) cream, (j) smart card, (k) cube, (l) disk, (m) duster, (n) eraser, (o) glue, (p) highlighter, (q) juice, (r) lemon, (s) mobile, (t) mouse, (u) screw-driver, (v) soap (w) spray, (x) sticky-notes .}
      \label{training}
      \vspace{-15pt}
   \end{figure}

Figure \ref{methodology} shows the block diagram of research methodology adopted to perform the \textit{human inspired robotic manipulation} tasks. The methodology starts with teaching basic grasping and manipulation primitives to the robot hand on the given training objects in Fig. \ref{training} by tele-operating it and recording the corresponding joint hand configurations. The Principle Component Analysis (PCA) is applied to all the individual tasks and the directions of highest data variance are selected as the predominant postural synergies. The postural synergies evolve over the duration of demonstration to obtain the corresponding synergistic trajectories. The given synergistic trajectories are approximated with Gaussian components using Gaussian Mixture Model (GMM) and the GMR generates the reference trajectory to reproduce the taught grasping and manipulation primitives. In order to generalize the learned synergy subspace to wider set of objects, the idea of Kernelized Movement Primitives (KMP) is exploited to deal with the environmental descriptors i.e, via-points and end-points for the new object's shape and size respectively. The kernel trick used in KMP preserves the probabilistic properties of synergy subspace such that it can be reused for grasping and manipulation of unknown objects. The geometrical features of the new objects are estimated from their point cloud using RANSAC algorithm with Euclidean clustering for object segmentation and the SVM classifier, trained on objects in Fig. \ref{training}, is used for object recognition. The pose of object is enumerated locally from the centroid of recognized object's point cloud. The estimated pose of object is then transformed into the respective synergistic values, which then act as environmental descriptors for the KMP to update its reference trajectory to adapt to the targeted object. Finally, the mapping from synergistic subspace to the joint space helps in performing the given tasks on the robot system.

\subsection{Kernelized Synergies Framework}

\begin{figure*}[t]
      \centering
      \includegraphics[height=3.5cm, width=16cm]{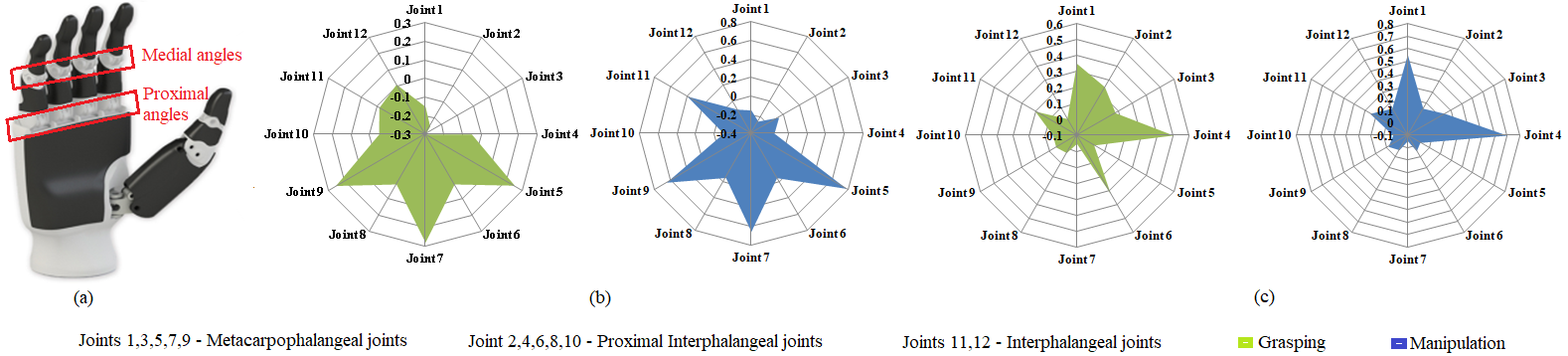}
      \caption{Graphical representation of computed synergistic subspace of INSPIRE robot hand, (a) represents the 6 DOF robot hand, (b) and (c) are the first two predominant synergies with respective grasping and manipulation components.}
      \label{synergies}
      \vspace{-15pt}
   \end{figure*}

The robot hand shown in Fig. \ref{synergies} (a) is tele-operated on the training objects in Fig. \ref{training} and the mean hand positions  ${\hat{\theta_k}}$=$\theta_k$-$q_0$, with $\theta_k$ and $q_0$ representing the current hand configuration and nominal hand posture respectively, are recorded. The mean hand positions are then concatenated into a row vector to construct a configuration matrix $C$=$[{\hat{\theta_1}}......{\hat{\theta_K}}]^{T}$ and the PCA is applied on $C$ to obtain the lower dimensional synergistic subspace $\hat{E}$ \cite{c14}.

The synergistic subspace is accessed by properly selecting the appropriate values of the synergy co-efficients $(e_i)={e_{g1}+e_{m1},e_{g2}+e_{m2},...e_{gn}+e_{mn}}$ for a given object to be grasped and manipulated and are determined by Eq. \ref{eqt_1}, where ${\hat{E}}^{\dagger}$ is the pseudo inverse of the synergy matrix

\begin{equation}
{e_i}={\hat{E}}^{\dagger}{({\theta_i}-{q_0})}
\label{eqt_1}
\end{equation}

The synergistic subspace of two predominant postural components, estimated using training objects in Fig. \ref{training}, for the robot hand in Fig. \ref{synergies} (a), is illustrated in Fig. \ref{synergies} (b and c). The computed subspace reveals that the first synergistic component regulates the medial and proximal joints of each finger while the second synergistic component controls the relative motion of index finger and thumb during grasping and manipulation. Further, such a relative motion of index finger and thumb helps in switching from one grasping position to other while manipulating an object.

The synergistic co-efficients in Eq. \ref{eqt_1} for each object in Fig. \ref{training} evolve over the duration of demonstration to obtain the corresponding synergistic trajectories $e(t)$. The synergistic trajectories are approximated with Gaussian components $e(t){\sim}{\sum_{n=1}^{N}}{\pi_n}{\mathcal{N}}({{\mu}_n},{{\Sigma}_{n}})$ using GMM, with ${\pi}_n$,${\Sigma}_n$,${\mu}_n$ representing the prior probability, covariance and mean of the $n^{th}$ Gaussian respectively and N being the total number of Gaussian components. Further, a reference synergistic trajectory, to be followed by the robot hand to reproduce the taught postures, is generated by conditioning the joint probability distribution of the GMM i.e, ${e_n}|t{\sim}{\mathcal{N}}({{\mu}_n},{{\Sigma}_n})$ using GMR.  

To generalize the learned synergy subspace and to adapt to new objects, the idea of KMP is exploited. For a new instance $t^{*}$ (new object), the expected mean and co-variance of the synergistic trajectory are computed using Eq. \ref{eqt_2} \cite{c15}, with $k^{*}$ being a kernel function i.e, ${{k}^{*}}=[{k}(t^{*},{t}_{1}),{k}(t^{*},{t}_{2}),.....{k}(t^{*},{t}_{N})]$, $K$ is the corresponding kernel matrix and $\lambda$ is a regularization parameter.

\begin{equation}
\begin{gathered}
{\E}(e(t^{*}))={k}^{*}({K}+{\lambda}{I})^{-1}{\mu} \\
\D(e(t^{*}))={\frac{N}{\lambda}}({k}({t}^{*},{t}^{*})-{k}^{*}({K}+{\lambda}{\Sigma})^{-1}k^{*T})
\end{gathered}
\label{eqt_2}
\end{equation}

The priorities $(\Upsilon)$ can be assigned to different grasping and manipulation tasks in the kernelized synergies framework and are formulated by Eq. \ref{eqt_3} \cite{c15}, which represents the product of M Gaussian distributions in the kernelized synergistic subspace with $m= 1,2,...M$.  

\begin{equation}
\mathcal{N}({\mu_n}^{T},{\Sigma_n}^{T}) {\propto}{\prod_{m=1}^{M}}\mathcal{N}({\hat{\mu}_{n,m}},{\hat{\Sigma}_{n,m}}/\Upsilon_{n,m})
\label{eqt_3}
\end{equation}

The compliance in the kernelized synergies for flexible interactions with the objects is introduced by the model of soft synergies ${\Delta}{q}={\hat{E}}{\Delta}{e}$ with ${\Delta}{q}={\Delta}{q_{ref}}-{C_h}{\Delta}{\tau}$, where $C_h$ represents the robot hand compliance matrix and $\tau$ is the vector of robot hand joint torques. Therefore, the grasp established with the kernelized synergies can be now defined by Eq. \ref{eqt_4} in terms of forces balancing the grasped object.

\begin{equation}
    {f_c}={G}^{\dagger}{\omega}+{\xi}{\Delta}{e}
\label{eqt_4}
\end{equation}

Where, ${f_c}$ represents the vector of contact forces between the object and robot hand fingers, ${G}^{\dagger}$ is the pseudo-inverse of grasp matrix, ${\omega}$ is the vector of external forces and wrenches applied to the grasped object, ${\xi}$ is the subspace of internal forces and ${\Delta}{e}$ is the change in kernelized synergistic co-efficients. The kernelized synergies co-efficients in Eq. \ref{eqt_4} are modulated until the steady state grasping posture is achieved and it is conditioned by the threshold on the robot hand motors' current to ensure the grasp stability.

\subsection{Object Detection and Synergy Mapping}

\begin{algorithm}
\SetAlgoLined
\Input{${P_c}\longrightarrow{\text{point cloud of scene}}$}   
\Output{${e_O}\longrightarrow{\text{synergistic values of object}}$}
 \While{$({p^{(i)}}\in{P_c})\longrightarrow{P_p}$}{
 $function (P_p)\longrightarrow(Inliers,Outliers)$\\
 $P_p := Ouliers$\\
 $function (P_p,C)\longrightarrow(O)$\;
  \eIf{$d(x,y) \leq {\epsilon}{|{MinPts}}$}{
   ${O}$\;
   }{
   $\phi{\longrightarrow}{\text{discard point}}$\;
  }
 }
\caption{Object Detection \& Synergy Mapping}
\ForEach{$(shape,color)\in{O}$}{
 $function (l,\psi) = SVM(O,model)$\;
  \uIf{${(shape,color)}\cap{(model)}\longrightarrow{\text{object detected}}$}{
   $\text{return}(l,\psi)$\;
   }
   \Else{
   $\phi{\longrightarrow}{\text{object not found}}$\;
   }
   ${P_o}^c = centroid(\psi)$ \\
   ${P_o}^b = transformation({P_o}^c)$ \\
   ${e_O} = mapping({P_o}^b)$
}
\end{algorithm}

   \begin{figure*}[t]
      \centering
      \includegraphics[height=3.5cm, width=18cm]{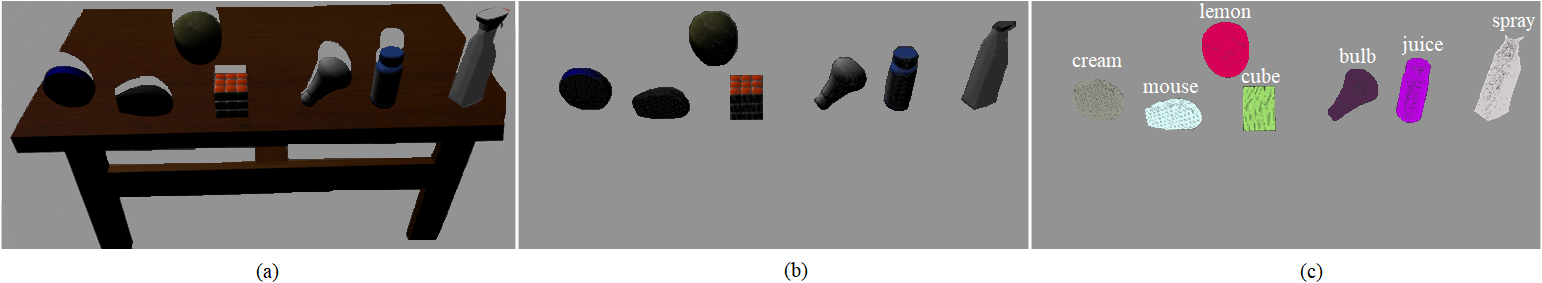}
      \caption{Object segmentation and recognition, (a) represents the raw point cloud of the scene, (b) illustrates the processed point cloud after filtering out the adversarial data points and removing the larger planar component (table) using RANSAC algorithm, (c) indicates the candidate objects being recognized and labeled by the trained SVM classifier.}
      \label{pose}
      \vspace{-15pt}
   \end{figure*}

The point cloud of the scene, consisted of objects placed on the table, is acquired using Intel RealSence D435 camera and is shown in Fig. \ref{pose} (a). The raw point cloud $P_c$ is filtered to remove the adversarial data points and is down-sampled to reduce the processing time. Further, RANSAC algorithm is used to separate the large planar component representing the table and the remaining points are grouped into different clusters that represent the candidate objects in the processed point cloud $P_p$ in Fig. \ref{pose} (b). The essential points in $P_p$, expressed in k-d tree, are grouped according to Euclidean clustering criteria, defined by Eq. \ref{eqt_5}. 

\begin{equation}
    O = {\int_{i=0}^{C}}N_{\epsilon}(x_i|MinPts):{\{y_i|d(x_i,y_i){\leq}{\epsilon}\}}dx
\label{eqt_5}
\end{equation}

Where, $\epsilon$ represent the radius of neighbour $N_\epsilon$ of a point, $MinPts$ is the minimum number of neighbouring points within $\epsilon$, point y can be grouped with point x, if the distance between them is less than or equal to $\epsilon$ otherwise discarded and C is the number of clusters in the processed point cloud.

   \begin{figure}[t]
      \centering
      \includegraphics[width=8.5cm]{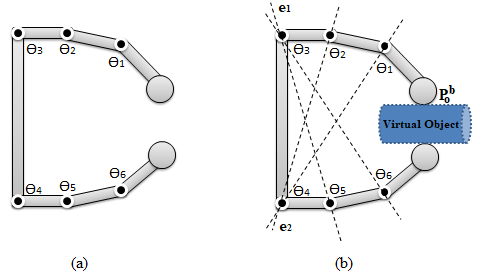}
      \caption{Object based synergistic mapping to robot hand. A virtual object similar to real one with estimated pose is assumed at the robot hand contact points and the corresponding synergistic components are extracted. The dashed lines represent the coordinated motion of robot hand within the synergistic subspace. }
      \label{mapping}
      \vspace{-15pt}
   \end{figure}

\begin{figure}[t]
      \centering
      \includegraphics[width=8.5cm]{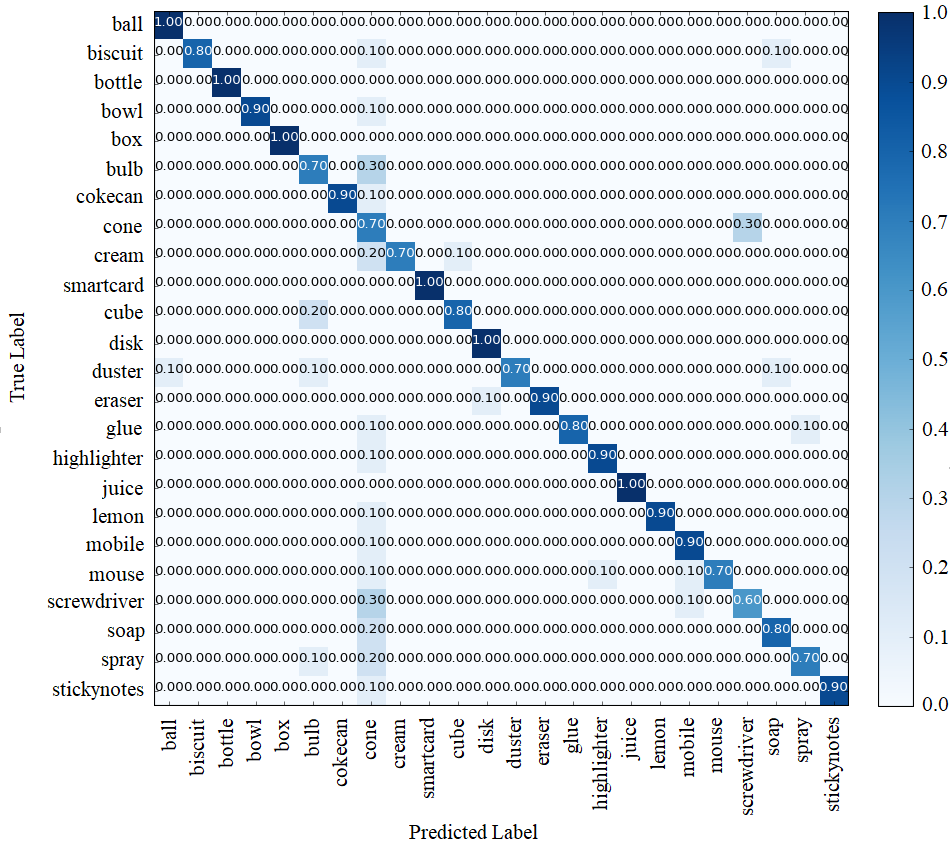}
      \caption{Normalized confusion matrix of trained SVM classifier.}
      \label{confusion}
      \vspace{-15pt}
   \end{figure}

The candidate objects found in term of clusters using Eq. \ref{eqt_5} are processed further to capture their shape and color characteristics using respective histogram features. Having segmented objects, the SVM classifier trained on objects in Fig. \ref{training} over 10 different orientations of each, successfully recognizes $(\psi)$ and labels $(l)$ the given objects, as shown in Fig. \ref{pose} (c). The performance of SVM classifier is assessed using confusion matrix in Fig. \ref{confusion} as an evaluation metric. It is evident from Fig. \ref{confusion} that trained classier has $84\%$ accuracy over the 240 instances in the data set.    

Such approach however ensures to grasp and manipulate novel objects but the approximations made with the given perception pipeline may introduce some errors in the object recognition i.e, sparse point cloud, occluded object, variational features etc. Thanks to the probablistic nature of kernelized synergies framework that uses Gaussian kernel having infinite Hilbert-Space, the system is robust against such perceptual errors. Finally, the pose of detected object is calculated locally by averaging all the data in the point cloud to determine the dimensions of their centroid. 

The geometrical features extracted with the given perception pipeline are in local camera frame and need to be transformed into robot base frame for precise targeted grasping and manipulation. The pose of the object in robot base frame is determined by using Eq. \ref{eqt_6}, where ${T_c}^m$ is the transformation between the camera frame and marker, ${T_m}^b$ is a fixed transformation between the marker and base frame, ${P_o}^c$ is the pose of object in the camera frame and ${P_o}^b$ is the pose of object in the robot base frame. 

   \begin{figure}[t]
      \centering
      \includegraphics[width=8.5cm]{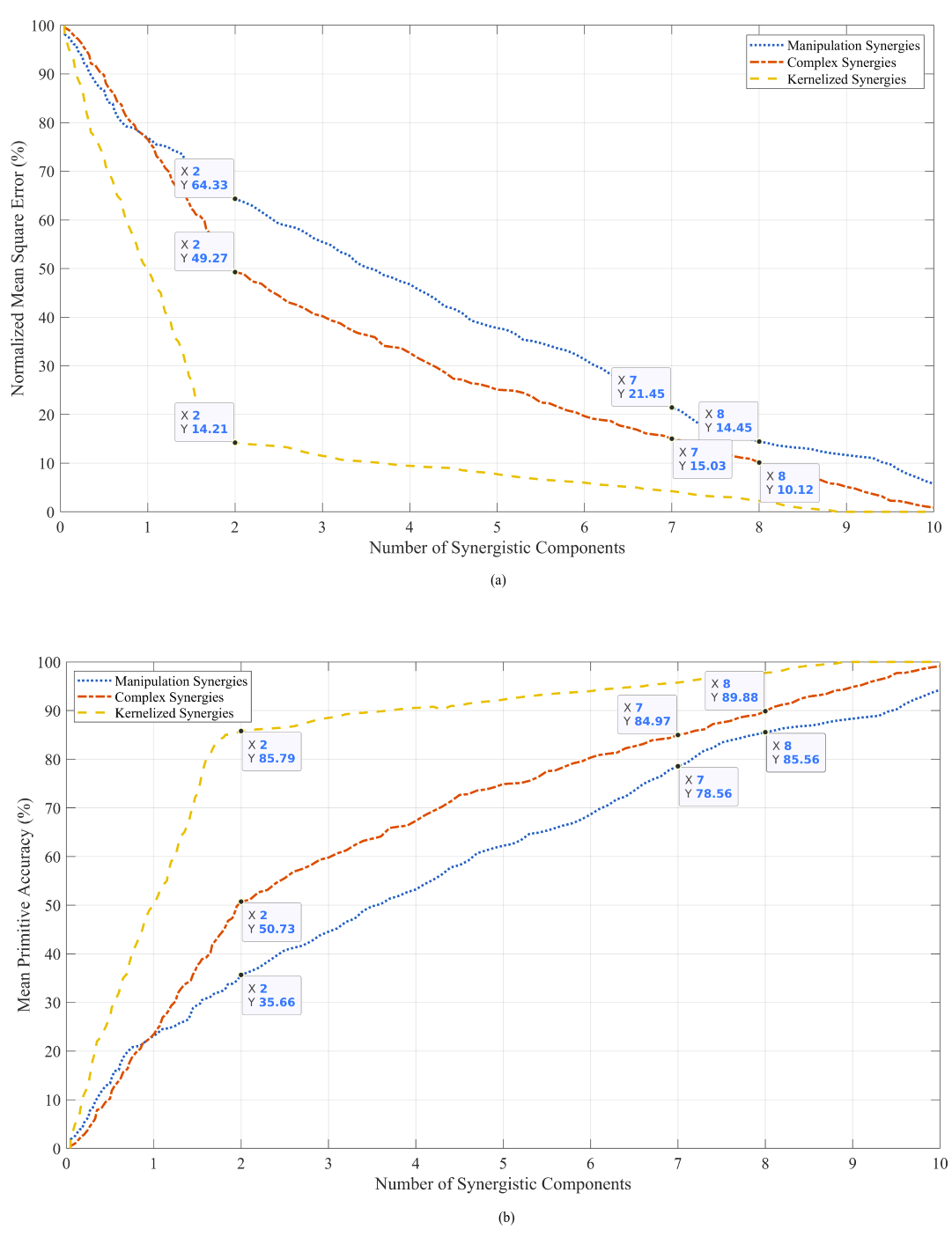}
      \caption{Performance evaluation of kernelized synergies framework against two different state of art techniques. Kernelized synergies achieve desired results in terms of normalized square error (a) and primitive accuracy (b) by exploiting the reduced number of synergistic components as compared to manipulation synergies and complex synergies approaches. }
      \label{comparison}
      \vspace{-15pt}
   \end{figure}

\begin{equation}
    {{P_o}^b} = {{T_m}^b}{{T_c}^m}{{P_o}^c}
\label{eqt_6}
\end{equation}

Further, the pose of object needs to be transformed into corresponding synergistic values such that it can update the via/end points of the KMP as described in Eq. \ref{eqt_2}. In order to do so, a virtual object of dimensions similar to the targeted shape is defined in the robot hand Cartesian space as shown in Fig. \ref{mapping}. The relationship between the robot hand Cartesian space and synergistic subspace is given by Eq. \ref{eqt_7}.

\begin{equation}
    {\dot{e_O}}={{A_m}^{\dagger}}{C_h}{\hat{E}^{\dagger}}{\dot{{P_o}^b}}
\label{eqt_7}
\end{equation}

Where, ${\dot{{P_o}^b}}$ is the robot hand Cartesian space velocity due to the object, ${C_h}$ is the hand compliance matrix, $A_m$ is the motion transfer matrix and ${\dagger}$ represents the pseudo inverse of respective quantities. The Euler integration at the sampling time of 1 ms is applied to Eq. \ref{eqt_7} to get the synergistic values of via/end points. Finally, the mapping from synergistic subspace to the joint space is required to execute the given task on the robot hand and is defined by Eq. \ref{eqt_8}, where $\theta_0$ represents the initial robot hand configuration. 

\begin{equation}
    {\theta}={\hat{E}}{e}+{\theta_0}
\label{eqt_8}
\end{equation}

The complete pipeline of object detection and synergy mapping is summarized in Algorithm 1.

\section{COMPARATIVE ANALYSIS}

   \begin{figure}[t]
      \centering
      \includegraphics[height=5cm, width=8.5cm]{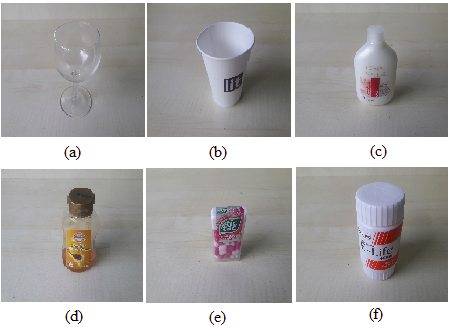}
      \caption{Set of test objects, (a) glass, (b) cup, (c) conditioner, (d) honey, (e) tic-tac, (f) food-supplement.}
      \label{test}
      \vspace{-15pt}
   \end{figure}

To better understand the performance and potential of kernelized synergies framework, it is compared with the other state of art techniques such as; manipulation synergies \cite{c16} and complex synergies \cite{c17} using two different evaluation metrics i.e, normalized square error (NSE) and primitive accuracy (PA), as shown in Fig. \ref{comparison}. The robot hand trained on objects in Fig. \ref{training}, is evaluated on test objects in Fig. \ref{test}. The robot hand performs the grasping (precision) and manipulation (rotation) of test objects and the difference between the robot hand configurations defined by the human subject(ground-truth) and established by the considered techniques is computed to define the given metrics in Fig. \ref{comparison}. The values of NSE and PA plotted in Fig. \ref{comparison}, are the average of all the hand configurations established over the test objects in Fig. \ref{test}.

As shown in Fig. \ref{comparison}, the manipulation synergies have high NSE and low PA because they are task specific and for any unknown object it requires to recompute the entire synergistic subspace on the new data set. Therefore, it completely changes the shape of whole synergy subspace and in particular the grasping properties of predominant postural synergies are not preserved. Such framework achieves the desired performance i.e, $15\%$ NSE and $85\%$ PA by exploiting almost $8$ synergistic components, which itself violates the motivation behind using synergies i.e, bringing simplicity. 

On the other hand, the complex synergies have relatively low NSE and high PA as compared to manipulation synergies and attain the desired accuracy but again at the cost of large number of synergistic components (although less than manipulation synergies) to reproduce the required hand configurations. However, this technique preserves the grasping properties of synergistic subspace and the manipulation tasks are executed by the additional synergistic components that are used to correct the corresponding hand configuration with respect to the grasping posture. Although, it does not require to re-compute the synergy subspace for every new entry in contrast to manipulation synergies but it increases the size of synergy subspace and is computationally inefficient. Further, the properties of grasping synergies are preserved locally and can not be generalized to even the different dimensions of same objects.  

It is evident from the results in Fig. \ref{comparison} that kernelized synergies have lowest NSE and highest PA as compared to other two techniques mainly due to two reasons; \textbf{(1)} the use of movement primitives to adapt to the dimensions of new geometrical shapes and \textbf{(2)} the application of kernel trick which preserves the probabilistic properties of computed synergy subspace globally such that it can be reused for grasping and manipulation of different objects. Further, this framework achieves the desired results by utilizing just first two predominant synergistic components.  

\section{RESULTS AND DISCUSSION}

To experimentally evaluate the performance of updated kernelized synergies framework, three different daily life tasks i.e, \textbf{(1)} mounting bulb on the socket, \textbf{(2)} squeezing lemon into the water, and \textbf{(3)} spraying cleanser on the board are performed.

\begin{figure}[t]
      \centering
      \includegraphics[width=8.5cm]{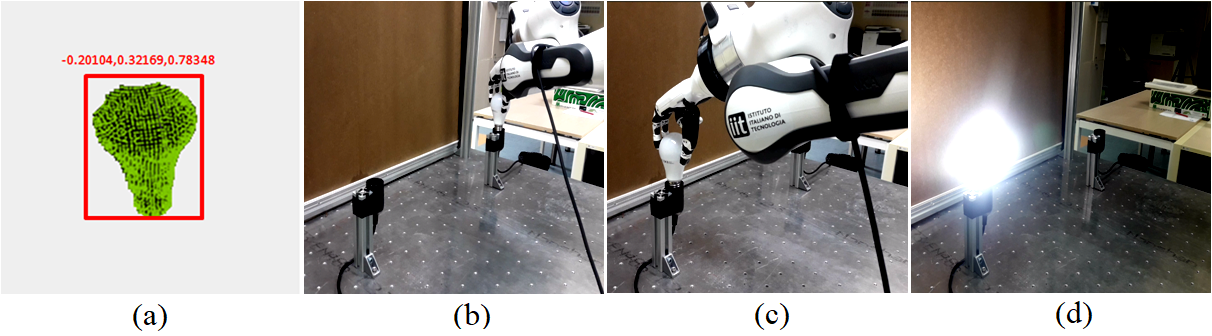}
      \caption{A robot mounting bulb on the socket, (a) is the estimated pose of bulb in the camera frame, (b) shows the robot hand grasping the bulb in its tripod posture, in (c) robot brings the bulb over the socket, (d) represents the bulb after being screwed into the socket.}
      \label{bulb}
      \vspace{-15pt}
   \end{figure}

In the first task, the robot hand performs bulb mounting task by exploiting the rotation primitives of its synergy subspace, similar to humans. For simplicity, the expired bulb has been already removed from the socket and the robot hand is asked to mount the new one. First of all, the eye-to-hand camera estimates the pose of the bulb using Eq. \ref{eqt_6} as shown in Fig. \ref{bulb} (a) and updates the kernelized synergies subspace according to Eq. \ref{eqt_8}. The updated values of kernelized synergies are computed using Eq. \ref{eqt_2} that help in mounting the bulb on the socket. The robot hand grasps the bulb in tripod posture at ${e_{g1}=-0.18, e_{g2}= 0.21}$ in Fig. \ref{bulb} (b) and places it on the socket using robot arm movements as in Fig. \ref{bulb} (c). The bulb is then rotated clockwise at $e_{m1}=-0.19$ to $-0.05$, $e_{m2}= 0.21$ to $0.37$ such that it is screwed into the socket as shown in Fig. \ref{bulb} (d). In this task, the force between the robot hand fingertips and the bulb is modulated approximately from 2.38 N to 3.57 N according to Eq. \ref{eqt_4}. It is due to the reason that the resistance between the bulb's rings and the socket increases during the final phase when it is about to mount on the socket and thus requires to apply more strength.

\begin{figure}[t]
      \centering
      \includegraphics[width=8.5cm]{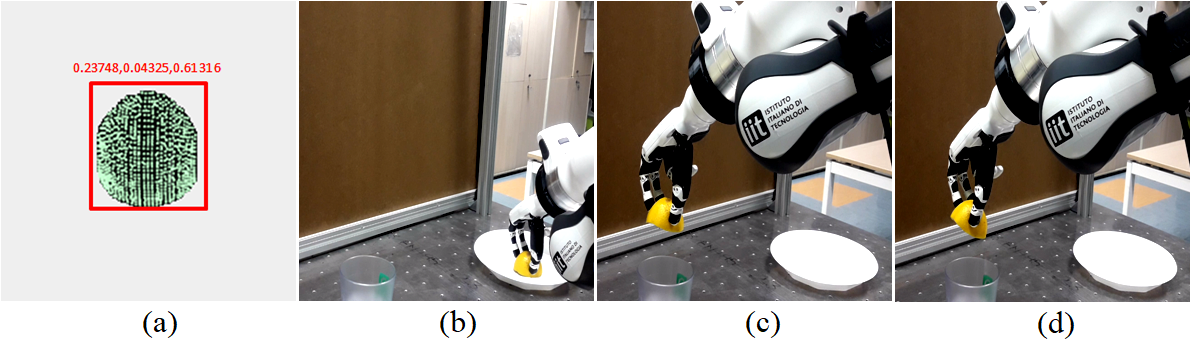}
      \caption{A robot squeezing lemon into the glass of water, (a) is the estimated pose of lemon, (b) shows the robot hand gripping lemon in its tripod posture, (c) illustrates the robot bringing lemon over the glass, (d) represents the robot hand squeezing lemon into glass.}
      \label{lemon}
      \vspace{-15pt}
   \end{figure}

During the second task, the robot hand utilizes the translation primitives of its synergy subspace to squeeze the lemon into the water to prepare the lemon juice. With the pose of lemon estimated as shown in Fig. \ref{lemon} (a), the robot hand grips it in a tripod posture in Fig. \ref{lemon} (b) at ${e_{g1}=0.12, e_{g2}= 0.32}$ and brings it over the glass using the robot arm movements in Fig. \ref{lemon} (c). The robot hand stretches in its index and middle fingers at $e_{m1}=0.13$ to $0.26$, $e_{m2}= 0.33$ to $0.47$ such that the lemon is pressed and its juice is sprinkled into the glass of water as shown in Fig. \ref{lemon} (d). Alike first task, the interaction force is modulated from 2.38 N to 4.16 N according to Eq. \ref{eqt_4}.

\begin{figure}[t]
      \centering
      \includegraphics[width=8.5cm]{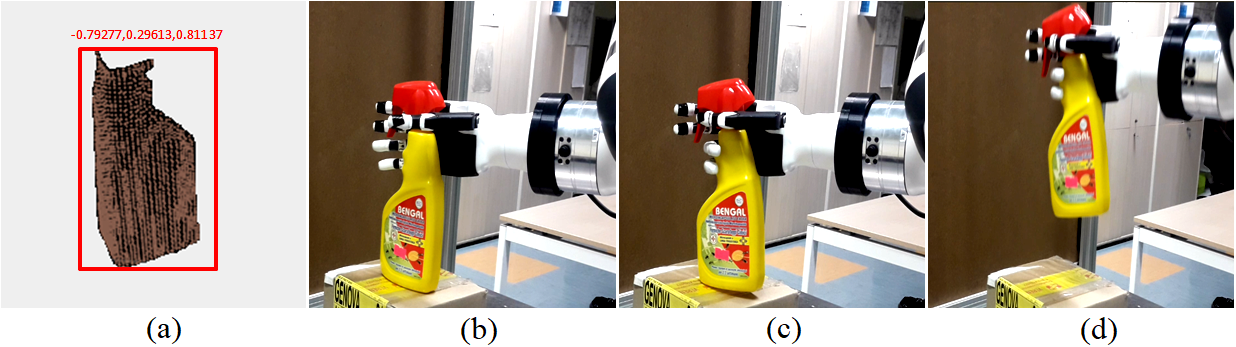}
      \caption{A robot spraying cleanser on the board, (a) is the estimated pose of spray in the camera frame, (b) is the pre-grasping posture of robot hand, (c) shows the robot hand holding spray with its thumb, little and ring fingers, (d) represents the robot hand pressing spray on the board with its index and middle fingers.}
      \label{spray}
      \vspace{-15pt}
   \end{figure}

For the third task, the robot hand takes an advantage of priority characteristics of kernelized synergies framework according to Eq. \ref{eqt_3}, to regulate its two parts independently. In this task, using the pose information of spray bottle obtained with object detection pipeline in Fig. \ref{spray} (a), the robot hand initially assumes the pre-grasping pose in Fig. \ref{spray} (b) and then grasps it in Fig. \ref{spray} (c) at ${e_{g1}=-0.11, e_{g2}= 0.38}$ with priority $\Upsilon = 0.5$ being assigned to each. The robot hand presses the spray within its tripod posture at $e_{m1}=-0.1$ to $0.14$, $e_{m2}= 0.39$ to $0.48$ while the index and ring fingers help in holding the spray as shown in Fig. \ref{spray} (d). The action of pressing the spray, results into showering of cleanser on the board. Similarly, in this task, the interaction force between the spray and robot hand is modulated from approximately 2.38 N to 4.76 N according to Eq. \ref{eqt_4}. Such priority characteristics of kernelized synergies framework are helpful in various multi-digit manipulation tasks such as; holding and typing on the smart phone, gripping and pressing the computer mouse, holding the receiver and dialing a number on the telephone and many others.

\section{CONCLUSIONS}

This research proposed to update the previously developed kernelized synergies framework with visual perception for autonomous grasping and manipulation of novel objects. The point cloud of objects in the scene was extracted with RGB-D camera and was pre-processed and later filtered using RANSAC algorithm to remove the largest planar component in the scene. The remaining points were grouped to represent the candidate objects, based on the Euclidean clustering segmentation criteria. The candidate objects in the filtered point cloud were recognized and labeled using the trained SVM classifier and the object pose was estimated locally by computing the centroid of each candidate object. The geometrical features of the detected objects were in image plane and the suitable transformations were applied to obtain the desired pose in the robot base frame for task execution. The use of RANSAC algorithm however introduced some approximation inaccuracies in the object recognition and pose estimation pipeline but due to the probabilistic nature of kernelized synergies, it was robust against the perceptual errors. Further, the performance of kernelized synergies framework was compared with the other state of art techniques and it was evident that kernelized synergies are effective, efficient and robust in performing various robust grasping and fine manipulation tasks. Finally, the updated framework was experimentally evaluated on three different daily life tasks i.e, mounting bulb on the socket, squeezing lemon into the water, and spraying cleanser on the board and the results confirm the task agnostic approach of the updated kernelized synergies framework.

For future work, the tactile information will be integrated into the kernelized synergies framework to do visuo-tactile dexterous manipulation. Further, the updated framework can also be extended to anthropomorphic dual arm-hand system for whole body manipulation in the cluttered environments.   

\bibliographystyle{IEEEtran}

\end{document}